\documentclass[letterpaper, 10 pt, conference]{ieeeconf}  

\IEEEoverridecommandlockouts                              

\usepackage{pifont}
\usepackage{xcolor}
\usepackage{makecell}

\newcommand{\cmark}{\textcolor{green}{\ding{51}}}
\newcommand{\xmark}{\textcolor{red}{\ding{55}}}

\usepackage{amsmath}

\usepackage{graphicx}

\usepackage{booktabs}
\usepackage{multirow}
\usepackage{amsfonts}
\usepackage{algorithm}
\usepackage{algpseudocode}
\usepackage{subcaption}
\usepackage{rotating}
\usepackage{array}
\newcolumntype{P}[1]{>{\centering\arraybackslash}p{#1}}

\let\labelindent\relax
\usepackage{enumitem}

\usepackage{balance}

\usepackage[compress]{cite}

\title{\LARGE \bf
Learning Locomotion on Complex Terrain for Quadrupedal Robots with Foot Position Maps and Stability Rewards
}

\author{Matthew Hwang$^{1}$, Yubin Liu$^{1}$, Ryo Hakoda$^{1}$ and Takeshi Oishi$^{1}$
\thanks{$^{1}$
        Authors are with the University of Tokyo, Japan
        {\tt\small email: mhwang@g.ecc.u-tokyo.ac.jp}}%
}

\begin{document}

\maketitle
\thispagestyle{empty}
\pagestyle{empty}

\begin{abstract}

Quadrupedal locomotion over complex terrain has been a long-standing research topic in robotics. 
While recent reinforcement learning-based locomotion methods
improve generalizability and foot-placement precision, they rely on implicit inference of foot positions from joint angles, lacking the explicit precision and stability guarantees of optimization-based approaches.
To address this, we introduce a foot position map integrated into the heightmap, and a dynamic locomotion-stability reward within an attention-based framework to achieve locomotion on complex terrain.
We validate our method extensively on terrains seen during training as well as out-of-domain (OOD) terrains. Our results demonstrate that the proposed method enables precise and stable movement, resulting in improved locomotion success rates on both in-domain and OOD terrains.


\end{abstract}

\section{INTRODUCTION}
Quadrupedal robots have the potential to traverse over complex and unstructured terrains that pose significant challenges for wheeled and crawler-based robots. To reliably deploy quadrupedal robots, robust locomotion strategies that can handle complex terrains are needed. Over the years classical optimization-based locomotion methods have been proposed to achieve locomotion on diverse terrain~\cite{Shkolnik2011little_dog,  Grandia2019FeedbackMPC, Ding_2021, Ruben2022perceptiveNMPC, Jenelten_2022}. More recent methods based on reinforcement learning (RL) propose to learn locomotion through trial and error~\cite{haarnoja2018learning, rudin2022learning, Lee_2020, Miki_2022}.


For locomotion over sparse terrain, much research has been conducted on optimization-based methods~\cite{9134750, tsounis2020deepgaitplanningcontrolquadrupedal, fahmi2022vital, Jenelten_2022}, RL-based methods~\cite{zhang2024learningagilelocomotionrisky, yu2025walkingterrainreconstructionlearning, wang2025beamdojolearningagilehumanoid, 11196002}, and hybrid methods~\cite{9779429, 10.1007/978-3-031-21090-7_31, Jenelten_2024}. 
Optimization-based methods exhibit high foot placement precision but lack robustness against real world noise and uncertainties. 
On the other hand, RL-based methods excel in robustness but lack foot placement precision. 
While DTC~\cite{Jenelten_2024} partially solves this issue, it has a complex architecture and is inherently constrained by the model-based planner utilized for foot planning~\cite{Jenelten_2022}. 
Recent research has utilized attention-based heightmap encoding~\cite{doi:10.1126/scirobotics.adv3604} to focus on important sections of the surrounding terrain. 
This increases generalizability of the trained policies on complex and sparse terrain. 

However, with AME~\cite{doi:10.1126/scirobotics.adv3604} the robot must implicitly infers where the feet are placed from the proprioception information. Furthermore, RL methods in general lack stability guarantees that optimization-based methods possess. Additionally, in some cases we found the robot learns to avoid dangerous terrain such as stepping stones when using local velocity tracking commands. 

In order to solve these issues, we propose a foot position map (footmap) in the same representation as the heightmap. 
Directly placing foot position information inside of the heightmap lets the attention mechanism know where the feet are positioned explicitly, which is crucial information when performing locomotion over sparse terrains. 
Furthermore, since the policy does not have to infer the foot positions from joint angles, our policy excels on out-of-domain (OOD) terrains. 
Furthermore, we introduce a dynamic stability reward which guides the policy to learn locomotion that respects the model-based stability criteria, leading to safer locomotion. 
We also introduce global velocity tracking to ensure that the robot cannot exploit the reward function by avoiding difficult terrain. 
These contributions lead to successful locomotion over complex terrain on in-domain and OOD terrains. 

\begin{figure}[t]
  \centering
  \large
  \includegraphics[width=\columnwidth]{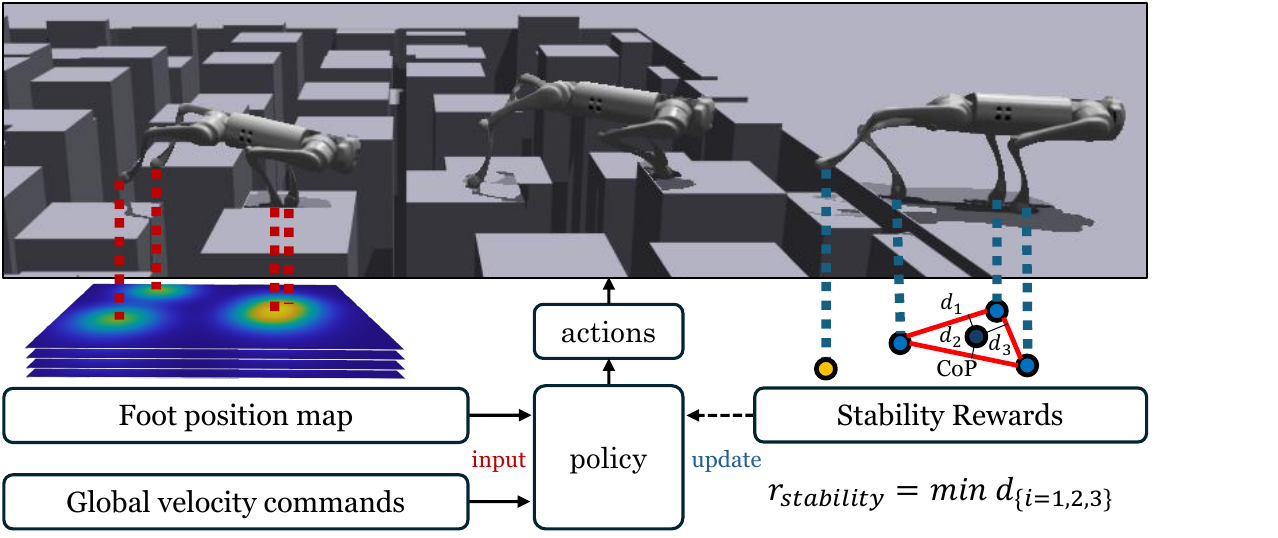}
  \caption{Overview of our method. With the proposed foot position map and stability reward, our policy achieves successful locomotion over complex terrain.}
  \vspace{-10pt}
  \label{fig:overview}
\end{figure}

\begin{figure*}[t]
  \centering
  \large
  \includegraphics[width=\linewidth]{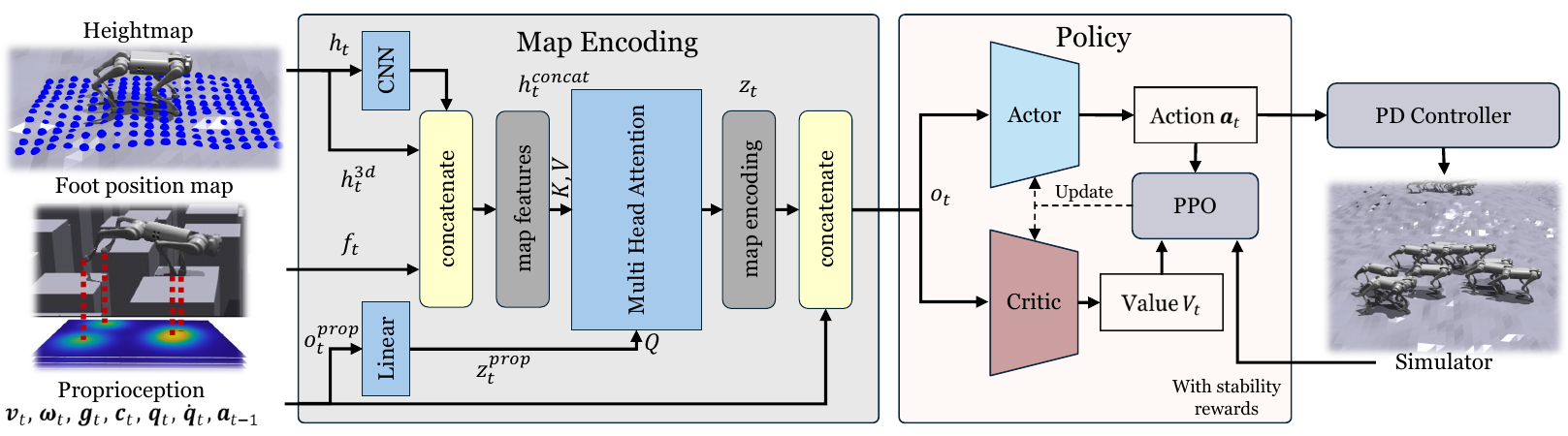}
  \caption{We propose a foot position map which is concatenated with the heightmap providing information of the foot positions relative to the terrain. During training, we add a stability reward  based on the CoP to guide the policy to prefer safer actions.}
  \label{fig:footmap_method}
\end{figure*}

\begin{figure}[t]
  \centering
  \large
  \includegraphics[width=\columnwidth]{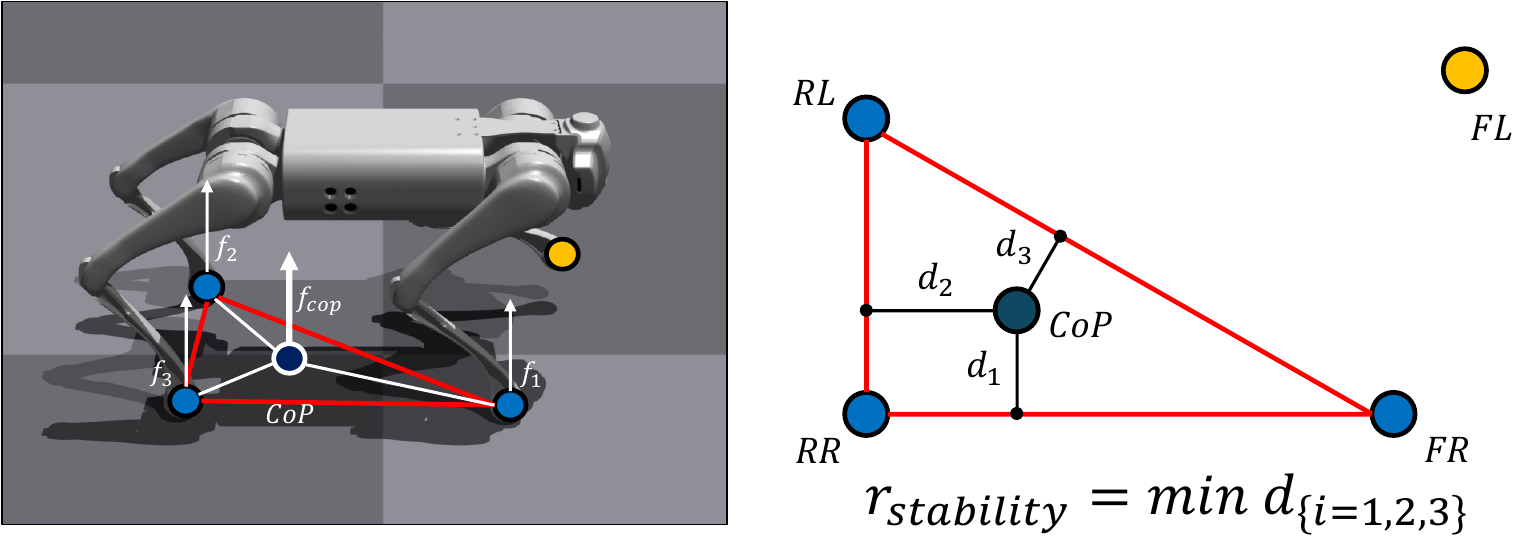}
  \caption{CoP-based stability rewards. We calculate the minimum distance of the \textbf{C}enter \textbf{o}f \textbf{P}ressure (CoP) to boundary of the support polygon as the stability reward.}
  \label{fig:cop_stability_rewards}
  \vspace{-10pt}
\end{figure}

The main contributions of this paper are three-fold:
\begin{enumerate}
    \item A novel foot position map embedded into the heightmap, providing direct knowledge of the relationship between the terrain and the foot positions;
    \item A dynamic stability reward that guides the policy to learn stable locomotion over complex terrain;
    \item Global velocity tracking, which prevents the robot from exploiting the reward function to avoid difficult terrain.
\end{enumerate}
We conduct extensive quantitative and qualitative evaluations across a wide range of terrains to validate our findings.
In addition, we evaluate our method with zero-shot sim-to-sim transfer to assess its applicability.

\section{RELATED WORKS}

\subsection{Locomotion over Sparse Terrain}


Much research has focused on locomotion over sparse footholds. DeepGait~\cite{tsounis2020deepgaitplanningcontrolquadrupedal} is comprised of a high level policy to output footholds and a low level policy which generates whole body motions to achieve these footholds, resulting on successful locomotion in simulation. DTC~\cite{Jenelten_2024} utilizes a state-of-the-art optimization-based foothold planner TAMOLS~\cite{Jenelten_2022} combined with RL for low level control by which they circumvent the problem of sparse reward signals on sparse terrain. However, the computational overhead for the optimization-based method is high. Moreover, the usage of an optimization-based foothold planner limits the robustness of the method compared to end-to-end learning-based methods. 

Recently, end-to-end methods have been researched in the context of quadrupedal and bipedal locomotion over sparse terrain. Zhang et al.~\cite{zhang2024learningagilelocomotionrisky} proposed a two stage curriculum learning method combined with a position-based goal definition~\cite{9981198} as an alternative to the standard velocity tracking formulation, achieving agile locomotion over sparse footholds. However, the MLP policy network overfits to the training terrain reducing generalizability. Yu et al.~\cite{yu2025walkingterrainreconstructionlearning} proposed to train a network to reconstruct the height map from depth images for locomotion over sparse foot holds incorporating the heightmap reconstruction network proposed in~\cite{10611621}. To avoid the problem of sparse reward signals on sparse terrain, Wang et al.~\cite{wang2025beamdojolearningagilehumanoid} proposed a soft-to-hard constraint training method. Most recently, Junzhe et al.~\cite{doi:10.1126/scirobotics.adv3604} proposed an attention-based heightmap encoder resulting in robust locomotion over sparse terrain with high generalizability.

\subsection{Stability in Locomotion}
In the context of legged locomotion the criteria of stability has been long researched. We first define a support polygon $\mathbf{P}$ which is the convex hull of the foot positions in contact with the ground as a series of continuous line segments $(\mathbf{l}_0, \dots,\mathbf{l}_n)$. The static stability of a robot can be calculated by whether the projected center of mass $\mathbf{CoM_{proj}}\in \mathbb{R}^{2}$ is within the robots support polygon $\mathbf{P}$~\cite{s20174911}. Keeping the projected CoM inside of the support polygon can be used as a constraint for the optimization of whole body movement during real time locomotion~\cite{8460731}. 

More recently, FT-Net~\cite{10305254} uses the distance of the CoM to the support polygon as a reward for locomotion with quadrupedal robots that have impaired joints. RLOC~\cite{9779429} utilize the Capture Point~\cite{4115602} for a stability reward and the CoM-CoP vector has been used for quadrupedal robots performing bipedal locomotion~\cite{Xiao2025}.


\section{METHODOLOGY}

In this Section we will introduce the three components of our proposed method and the training pipeline. First, we describe the proposed foot position map and the heightmap encoding in detail. Next we introduce the stability rewards. Finally, we explain the global velocity tracking and compare it to other command types. The rest of the section is dedicated to discussing the details of the observation and the action space of the policy and the reward functions.

\subsection{Foot Position Map Generation}
In our method we propose an explicit representation of foot positions integrated it into the heightmap. 
With the proposed foot position map the attention mechanism~\cite{doi:10.1126/scirobotics.adv3604} can attribute weights based on the foot position information without inferring the foot positions from joint encoder information. 
The heightmap is a $1.0~m \times1.6~m$ 2d gridmap of the heights surrounding the base of the robot sampled every $0.1~m$ with dimensions $H\times W = [17,11]$.
We implement the foot position map as a discrete 2d map with the same dimensions as the heightmap with one channel for each foot resulting in a tensor with dimensions ${H \times W \times 4}$. 
We represent the foot position map as a discretized gaussian distribution $\boldsymbol{f}_t \in \mathbb{R}^{H \times W \times 4} $ at time $t$ as given below: 
%
\begin{eqnarray}
\boldsymbol{f}_t \!\!\!\!&=&\!\!\!\! \lbrack \boldsymbol{f}_1(\mathbf{x}, t), \boldsymbol{f}_2(\mathbf{x}, t), \boldsymbol{f}_3(\mathbf{x}, t), \boldsymbol{f}_4(\mathbf{x}, t)\rbrack^\top, \\
\boldsymbol{f}_k(\mathbf{x}, t) \!\!\!\!&=&\!\!\!\! w \cdot \exp\left(-\frac{d_k^2(\mathbf{x})}{2\sigma^2}\right), k=1,...,4,
\label{eqn:footmap}
\end{eqnarray}
%
where $d_k$ is the distance from the sampled coordinates to the corresponding foot $k$. 
$\mathbf{x} \in \{\mathbf{x}_{ij}\}_{i=1,j=1}^{H,W}$ are the xy coordinates that we sample, where $k$ is the index of each foot and $i \times j$ is the heightmap dimension. $w$ is the foot position map weight, and $\sigma$ is the standard deviation. In our experiments we set $w$ to 10.0 and $\sigma$ to 0.1.

For deployment, we perform Forward Kinematics to calculate the foot position from the current joint angles. 

\subsection{Heightmap Encoding}
We propose to apply attention to the concatenated heightmap information which is a foot position map concatenated with the heightmap 3d coordinates and features, which we will describe in this section. The heightmap $\boldsymbol{h}_t \in \mathbb{R}^{H \times W}$ is first processed by a CNN to produce the heightmap features $\boldsymbol{z}^{heightmap}_t  \in \mathbb{R}^{H \times W \times (d - 7)}$. The CNN has two layers with a stride of 5 and padding. The first layer has dimensions of 16 and the second layer has dimensions of $d-7$. $d=64$ is the input dimensions of the MHA network:
%
\begin{equation}
\boldsymbol{z}^{heightmap}_t = \text{CNN}(\boldsymbol{h}_{t}).
\label{eqn:cnn}
\end{equation}
Then we concatenate the heightmap 3d coordinates $\boldsymbol{h}^{3d}_t \in \mathbb{R}^{H \times W \times 3}$, the heightmap features $\boldsymbol{z}^{heightmap}_t$ and the foot position map $\boldsymbol{f}_t$ to create the concatenated heightmap features $\boldsymbol{h}^{concat}_t  \in \mathbb{R}^{H \times W \times d}$:
\begin{equation}
\boldsymbol{h}^{concat}_t = \text{concat}(\boldsymbol{h}^{3d}_t, \boldsymbol{z}^{heightmap}_t, \boldsymbol{f}_t).
\label{eqn:concat}
\end{equation}
The proprioception encoding $\boldsymbol{z}^{prop}_t \in \mathbb{R}^{d}$ is created from the linear projection of the proprioception observation $\boldsymbol{o}^{prop}_t = [\boldsymbol{v}_t, \boldsymbol{\omega}_t, \mathbf{g}_t, \mathbf{c}_t, \mathbf{q}_t, \dot{\mathbf{q}}_t, \mathbf{a}_{t-1}]^{\mathrm{T}}$. With linear velocity $\boldsymbol{v}_t \in \mathbb{R}^{3}$, angular velocity $\boldsymbol{\omega}_t \in \mathbb{R}^{3}$, the gravity vector $\mathbf{g}_t \in \mathbb{R}^{3} $, linear and angular velocity commands $\mathbf{c}_t \in \mathbb{R}^{3}$, joint angles $\mathbf{q}_t \in \mathbb{R}^{12}$,joint velocities $\dot{\mathbf{q}}_t \in \mathbb{R}^{12}$, and previous actions $\mathbf{a}_{t-1}\in \mathbb{R}^{12}$:

\begin{equation}
\boldsymbol{z}^{prop}_t = \text{Linear}(\boldsymbol{o}^{prop}_t).
\label{eqn:prop_linear}
\end{equation}
Finally, we apply Multi Head Attention(MHA)~\cite{vaswani2023attentionneed} to the concatenated heightmap features $\boldsymbol{h}^{concat}_t$ with the query as $\boldsymbol{z}^{prop}_t$ to produce the heightmap encoding $\boldsymbol{z}_{t}$:
\begin{equation}
\boldsymbol{z}_{t} = \text{MHA}(K, V=\boldsymbol{h}^{concat}_t, Q=\boldsymbol{z}^{prop}_t).
\label{eqn:att}
\end{equation}

\subsection{Stability Rewards}



In our approach, we employ the dynamic stability criteria based on the center of pressure (CoP)~\cite{doi:10.1142/S0219843604000083} for the reward function. 
As shown in Fig.~\ref{fig:cop_stability_rewards}, we calculate the stability reward $r_{stability}$ from the minimum distance $d$ of CoP to $n$ sides of the support polygon:
\begin{equation}
r_{stability} = \min d_{i=1, ..., n}.
\label{eqn:dynamic_stability_reward}
\end{equation}

This stability reward guides the policy to maintain the CoP near the middle of the support polygon therefore enhancing the stability during locomotion. In the rare case that the CoP is outside of the support polygon (\textit{e.g.} when a leg is caught in the terrain and negative pressure is applied) we apply a negative penalty. 

\subsection{Global Velocity Tracking}
Local velocity tracking is commonly utilized when learning quadrupedal locomotion. With local velocity tracking, we randomly sample command velocities in the robot's local coordinates. This means that if the robot turns in any direction, as long as it moves at the given velocity in its local coordinates the robot can obtain high rewards. This setup generally works well but can be exploited by the robot on certain terrains such as stepping stones. We found that in some cases the robot learned to avoid the stepping stones and move in a circle inside of the starting platform. Some research propose to use position commands~\cite{9981198} that prevent the robot from avoiding difficult terrain. However, training becomes more difficult and accurate velocity tracking is challenging with this method.

In contrast, we propose a simple global velocity tracking method. With global velocity tracking, we first sample a random velocity in the global coordinates. Then, we apply a transformation from the global coordinates to the local coordinates of the robot. This way the robot cannot gain rewards by running away, mitigating the reward exploitation issue. With $\boldsymbol{R}^{T}$ as the Rotation matrix from global to local coordinates we calculate the local velocity commands $\boldsymbol{c}_t$ as, 

\begin{equation}
  \boldsymbol{c}_t =\begin{bmatrix}\boldsymbol{R}^{T}(\boldsymbol{v}^\text{global}_{cmd})\\ \omega_{cmd}\end{bmatrix} 
  \label{eqn:global_vel_tracking}
\end{equation}

consequently,  $\boldsymbol{v}^\text{local}_{cmd} = \boldsymbol{R}^{T}(\boldsymbol{v}^\text{global}_{cmd})$.

\begin{table}[t]
\centering
\small
\setlength{\tabcolsep}{3pt}  
\begin{tabular}{r|P{6em}P{6em}P{6em}}
\toprule
Method & No Reward Hacking & Velocity Accuracy & Simple Training \\
\toprule
Local Velocity & \xmark & \cmark & \cmark \\
Position & \cmark & \xmark & \xmark \\
Global Velocity & \cmark & \cmark & \cmark \\
\bottomrule
\end{tabular}
\caption{Command type comparison. Global velocity tracking maintains velocity tracking accuracy and simplicity of training while preventing reward exploitation.}
\label{tab:command_types}
\vspace{-10pt}
\end{table}

\subsection{Observation Space}
The policy network $\pi_{\phi}(\mathbf{a}_t|\mathbf{o}_t)$ takes $\mathbf{o}_t \in \mathbb{R}^{112}$ as the input and outputs actions $\mathbf{a}_t \in \mathbb{R}^{12}$. This policy is a neural network with parameters $\phi$.

The observation $\mathbf{o}_t$ is the proprioception observation data $\mathbf{o}^{prop}_t \in \mathbb{R}^{48}$ and the heightmap encoding $\mathbf{z}_t \in \mathbb{R}^{64}$:

\begin{equation}
  \mathbf{o}_t=[\boldsymbol{o}^{prop}_t, \mathbf{z}_{t}]^{\mathrm{T}}.
  \label{eqn:observation_vector}
\end{equation}

\subsection{Action Space}
In this research we employ a quadrupedal robot with 3 actuators per leg as the experimental platform. Consequently, the action space dimension of the policy network corresponds to the robot's actuator configuration, which consists of 12 degrees of freedom.

We define the actions as the difference from the default joint angles $\mathbf{q}_\text{default} \in \mathbb{R}^{12}$. Therefore, the final joint target angles $\mathbf{q}^\text{target}_t\in \mathbb{R}^{12}$ are calculated as,

\begin{equation}
  \mathbf{q}^\text{target}_t = \mathbf{q}_\text{default} + \alpha\mathbf{a}_t.
  \label{eqn:q_des}
\end{equation}

With $\alpha$ being a constant multiplier set to 0.25.
The final actuator torques $\boldsymbol{\tau} \in \mathbb{R}^{12}$ are calculated from a PD controller with $K_p=40$ and $K_d =1.0$:

\begin{equation}
  \boldsymbol{\tau}_t= K_p (\mathbf{q}^\text{target}_t - \mathbf{q}_t ) - K_d \dot{\mathbf{q}_t}.
  \label{eqn:torque}
\end{equation}

\subsection{Reward Functions}
We train a locomotion policy which follows velocity commands. The reward functions can be categorized into (1) Velocity command tracking rewards, (2) Regularization rewards to suppress dangerous movements and (3) Our proposed stability reward. The details of the reward functions can be found in Table~\ref{table:rewards}. $\boldsymbol{v}^{local}_{cmd}$, $\boldsymbol{v}^{local}_{b}$ are the command linear velocity and the actual linear velocity. $\omega_{cmd}$, $\omega_{yaw}$ are the command angular velocity and the actual angular velocity. $z_\text{target}$, $z$ are the target height of the robot and the actual height of the robot. $f$ is the contact force of the feet of the robot. $n_\text{collision}$, $n_{\|\mathbf{f}_{xy}\| > 5\|\mathbf{f}_{z}\|}$ are the number of collisions as well as the amount of times the collision force of the foot was greater in the horizontal direction in comparison with the force of the vertical direction respectively. The total reward $r_t$ at time step $t$ is calculated as:
\begin{equation}
  r_t(\mathbf{o}_t,\mathbf{a}_t) = \sum_i r_i w_i ,
  \label{eqn:reward_sum}
\end{equation}
where $i$ denotes different items listed in Table \ref{table:rewards}, and $w_i$ is the weight for the corresponding reward item.

\begin{table}[tb]\centering
\caption{Definition of rewards $\phi(x):=\exp(-\frac{||x||^2}{0.5})$.}
\tabcolsep=0.11cm
\begin{tabular}{crcl}
    \toprule
     & Reward Terms & Definition & Weight\\
    \toprule
     \multirow{2}{*}{(1)} & Linear Velocity Tracking & $\phi(\boldsymbol{v}^{local}_{cmd} - \boldsymbol{v}^{local}_{b})$ & $1.5$ \\
     & Angular Velocity Tracking & $\phi(\omega_{cmd} - \omega_{yaw})$ & $0.5$\\
    \hline
     \multirow{10}{*}{(2)} & Z Linear Velocity Penalty & $-\boldsymbol{v}_{z}^2$ & $1.0$ \\
     & Angular Velocity Penalty & $-||\boldsymbol{\omega}_{}||^2$ & $0.05$\\
     & Joint Torques & $-||\boldsymbol{\tau}_j||^2$ & $0.0001$\\
     & Joint Acceleration & $-||\dot{\mathbf{q}}_j||^2$ & $2.5e^{-7}$\\
     & Base Height Tracking & $-||z_\text{target} - z||^2$ & $1.0$\\
     & Action Rate & $-||\dot{\mathbf{a}}_j||^2$ & $0.03$ \\
     & Collisions & $-n_\text{collision}$ & $1.0$\\
     & Stumble & $-n_ {\|\mathbf{f}_{xy}\| > 5\|\mathbf{f}_{z}\|}$ & $0.1$\\
     & Joint Error & $-||\mathbf{q}^\text{default}_j - \mathbf{q}_j||^2$ & $0.04$\\
    \hline
     (3) & Stability & $r_\text{stability}$ & 1.0 \\ 
    \bottomrule\\
\end{tabular}
\label{table:rewards}
\vspace{-10pt}
\end{table}

\section{EXPERIMENTS}
\subsection{Experiment Setting}

\begin{figure}[tb]
    \centering
    \large
    \includegraphics[width=\columnwidth]{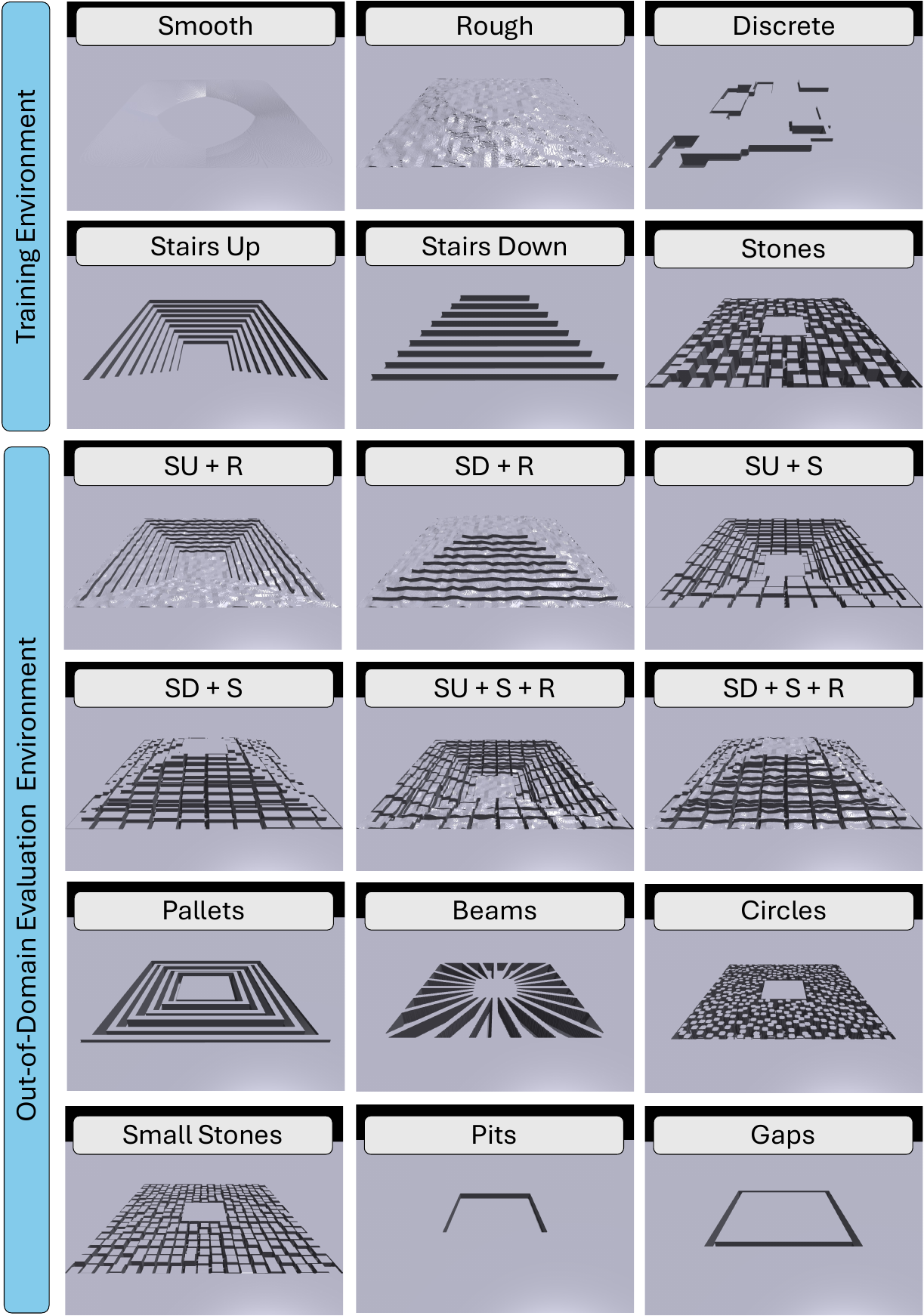}
    \caption{The training environment consists of smooth, rough, stairs up, stairs down, discrete and stones terrain. The OOD evaluation terrains include novel combinations of stones (S), rough (R), stairsup (SU) and stairsdown (SD) terrains as well as beams, pallets, circles, small stones, pits and gaps.}
    \label{fig:train_eval_env}
\end{figure}

\begin{table}[tb]
\footnotesize
\centering
\caption{Parameters of the stone terrain}
\label{tab:stones_params}
\begin{tabular}{c|ccccc}
\hline
\textbf{\shortstack{Difficulty \\ Level}} & \textbf{\shortstack{Stone Size \\ cm}} & \textbf{\shortstack{Stone Gap \\ cm}} & \textbf{\shortstack{Max. Shift \\ cm}} & \textbf{\shortstack{Max. Height \\ cm}} \\ \hline
0 & 92 & 2.5  & 0.0  & 1.0\\ 
$\cdots$&$\cdots$&$\cdots$&$\cdots$&$\cdots$\\
9 & 40 & 20 & 10 & 10\\
\hline
\end{tabular}
\vspace{-10pt}
\end{table}

\begin{figure*}[t]
    \centering
    \large
    \hspace{-15pt}
    \includegraphics[width=0.98\textwidth]{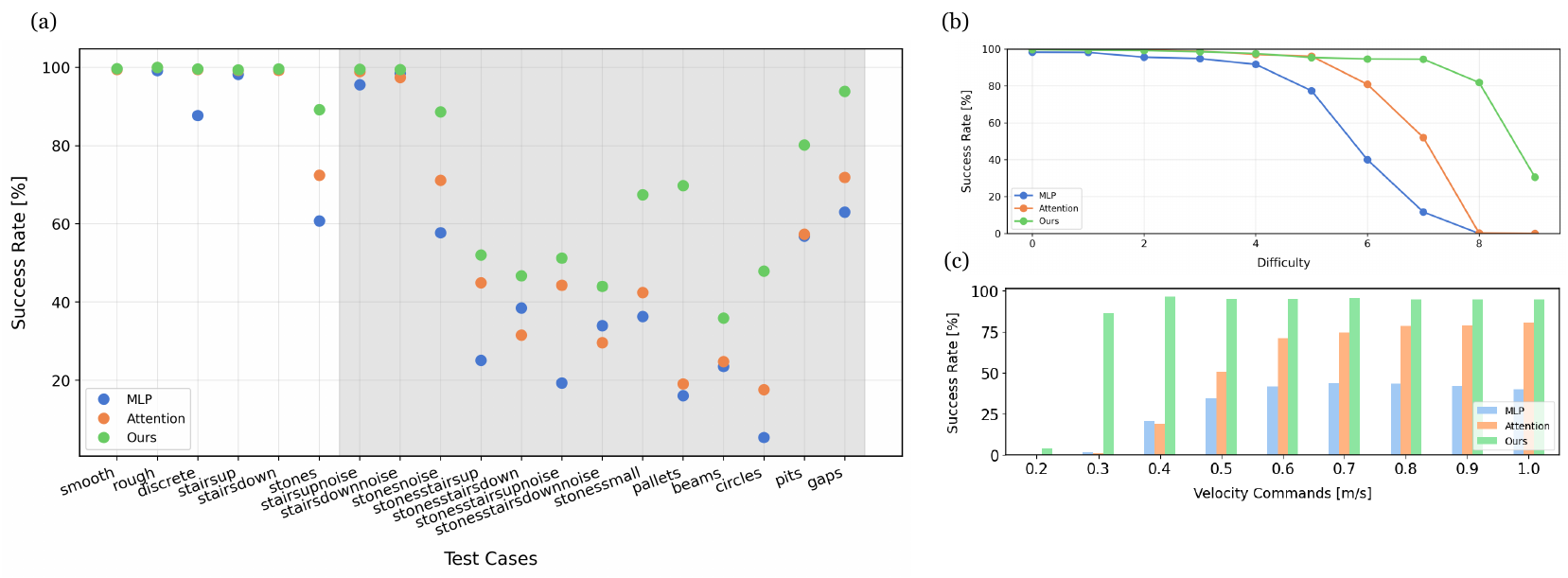}
    \caption{The success rate results. (a) Our proposed method produces higher success rates across the board for all terrains. (b) We observe an improvement at higher difficulty levels.  (c) We observe a higher success rate especially at lower velocities. }
    \label{fig:quantitative_results}
\end{figure*}


\subsubsection{Training Environment}
The training environment consists of a curriculum setup of 6 terrains: smooth, rough, discrete, stairs up, stairs down and stones as depicted in~\ref{fig:train_eval_env}. 


The stones terrain is implemented with the parameters specified in \ref{tab:stones_params} with varying stone sizes, stone gaps, random shifts and random heights. 


\subsubsection{Training Details}
we trained 4096 environmnets in parallel on a desktop PC with a single NVIDIA RTX4090. Training for our method took around 25 hours to complete. We use PPO~\cite{PPO} as the training algorithm with the legged gym environment~\cite{rudin2022learning} based on Isaac Gym~\cite{makoviychuk2021isaac}.  

\subsubsection{Evaluation Environment}
In our experiments we create a large variety of OOD terrains which are not encountered during training for testing as shown in \ref{fig:train_eval_env}. We introduce terrains from combinations of stepping-stones (stones:S), stairs(stairsup, stairsdown:SU, SD) and rough(R) terrain. Additionally we introduce 6 more terrains including pallets, beams, circles, smallstones, pits and gaps. 

\subsubsection{Evaluation Details}
For evaluation we sequentially generate a terrain from difficulty levels 0 to 9 and evaluate each individually. During evaluation for Fig.~\ref{fig:quantitative_results}-(a) and Fig.~\ref{fig:quantitative_results}-(b), we command the robot to move forward at 1.0 m/s for 20s. If the robots base does not make contact with the terrain and the robot moves more than 4m away from the origin, we calculate that trajectory as a success. We obtain the individual success rate of 1000 robots in simulation and calculate the mean as the final success rate at that difficulty level. 

For the success rate evaluation at different velocities of Fig.~\ref{fig:quantitative_results}-(c), we evaluate locomotion at command velocities from 0.1m/s to 1.0m/s. If the robots base does not make contact with the terrain and the robot moves more than half the expected distance based on the velocity command we calculate that trajectory as a success.

\subsubsection{Comparison Methods}
We choose 4 baselines to evaluate along side our proposed method. 

\begin{enumerate}[label=(\arabic*)]
  \item \textbf{Ours}: This is our proposed method with the foot position map, stability rewards and global velocity tracking.
  
  \item \textbf{Attention}: We utilize the same architecture from~\cite{doi:10.1126/scirobotics.adv3604}, which applies attention to the heightmap, filtering out unimportant heightmap information.
  
  \item \textbf{Transformer}: We implement a Transformer-based Encoder similar to LocoFormer~\cite{yang2022learning}. In the original paper, a depth image is used for exteroception. To have a fair comparison, we input a heightmap instead of the depth image.
  
  \item \textbf{CNN}: We implement a simple CNN-based heightmap encoder with 2 convolutional layers and a final linear layer.
  
  \item \textbf{MLP}: This is a simple MLP-based policy with a flattened heightmap as the input. 
\end{enumerate}

\begin{table}[tb]
\footnotesize
\centering
\caption{Results of comparison methods. We osbsere a constant improvement of the success rate, the survival rate tracking error and power consumption for our proposed method.}
\label{tab:overall_performance}
\centering
\renewcommand{\arraystretch}{1.25}
\setlength{\tabcolsep}{3pt}
\begin{tabular}
{r|P{3em}P{3em}P{3em}P{3em}P{3em}}
\hline
\textbf{Method} & MLP & CNN & Transf. \cite{yang2022learning} & Attn \cite{doi:10.1126/scirobotics.adv3604} & Ours \\
\hline
\textbf{Success rate} [\%]~$\uparrow$ & 58.7 & \underline{69.8} & 56.8 & 64.2 & \textbf{77.0} \\
\textbf{Survival rate} [\%]~$\uparrow$ & 62.1 & \underline{70.8} & 57.0 & 68.7 & \textbf{78.5} \\
\textbf{Tracking error} [$m/s$]~$\downarrow$ & 0.268 & 0.249 & \underline{0.229} & 0.243 & \textbf{0.169} \\
\textbf{Power} {[}W{]}~$\downarrow$ & 69.3 & \underline{60.2} & 78.7 & \underline{60.2} & \textbf{58.5} \\
\hline
\end{tabular}
\end{table}

\begin{figure*}[t]
    \centering
    \large
    \includegraphics[width=0.95\textwidth]{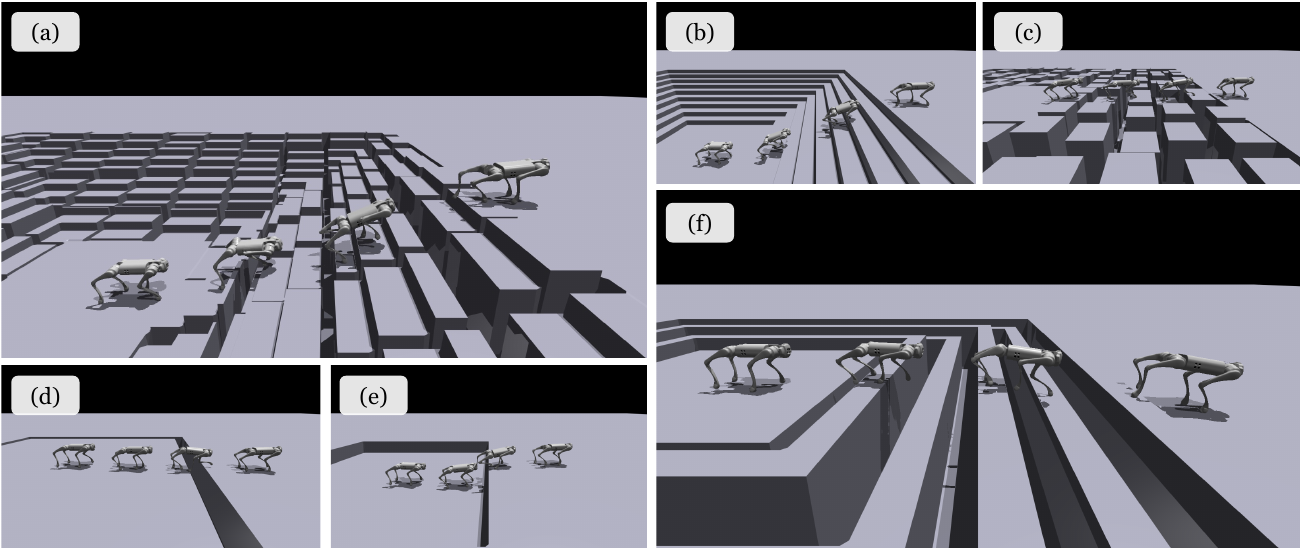}
    \caption{Qualitative locomotion results of our proposed method on different terrains. (a) stones + stairs up, (b) stairs up, (c) stones, (d) gaps, (e) pits, (f) pallets. }
    \label{fig:qualitative_locomotion_results}
    \vspace{-10pt}
\end{figure*}

\begin{figure}[tb]
    \centering
    \large
    \hspace{-13pt}
    \includegraphics[width=0.98\columnwidth]{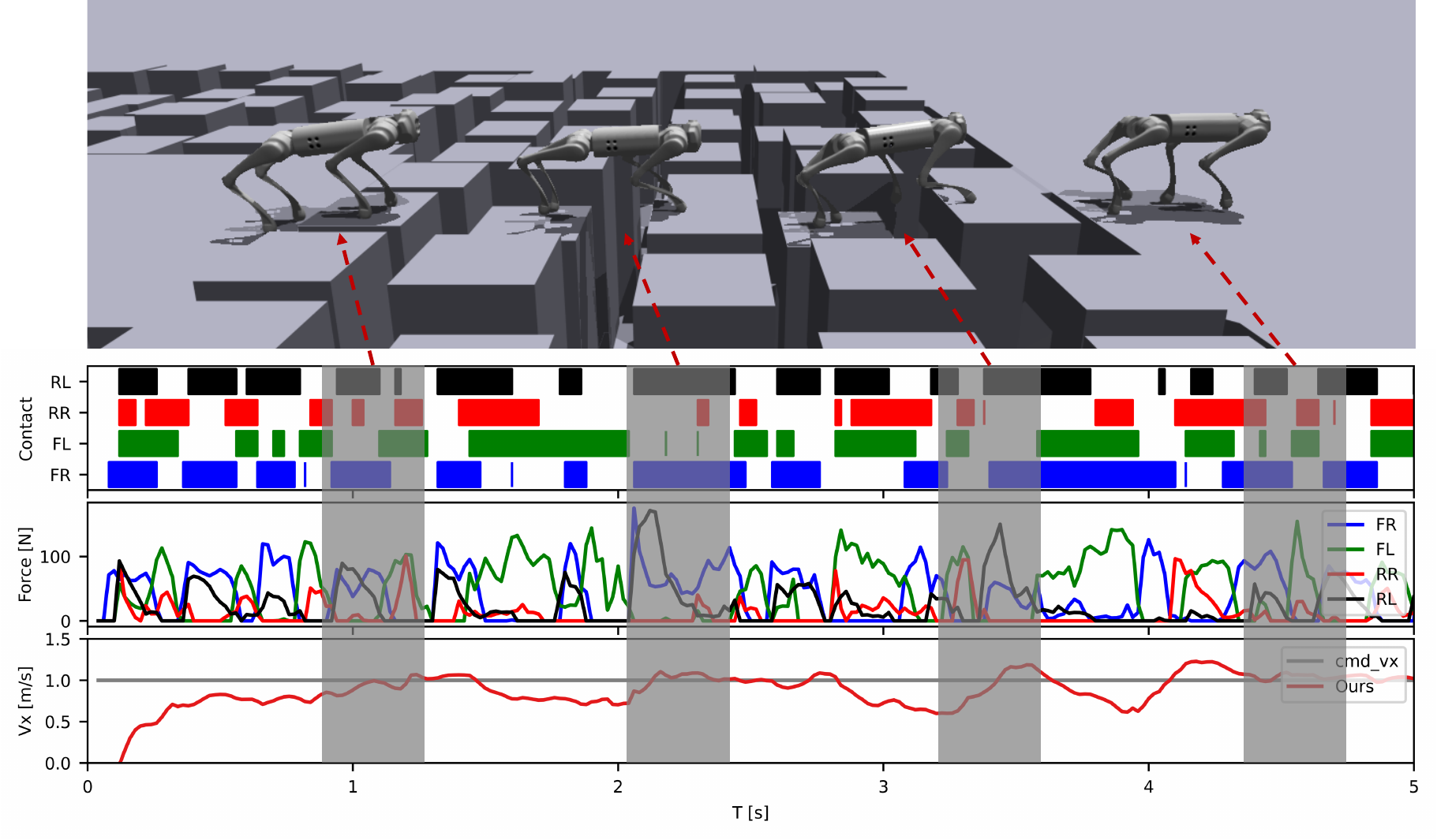}
    \caption{Qualitative locomotion results of our method on stones. (FR: Front Right, FL: Front Left, RR: Rear Right, RL: Rear Left)}
    \label{fig:locomotion_result_stones_ours}
\end{figure}

\section{RESULTS IN SIMULATION}
\subsection{Quantitative results}

\subsubsection{Success rate per terrains}
Fig. \ref{fig:quantitative_results}-(a) depicts the success rate of each method when evaluated on a diverse set of terrains. The columns that are highlighted with a gray color are the OOD terrains that we have added for the evaluation. Compared to the Attention policy our proposed method yields near 10\% increase. We observe our proposed method constantly outperforms the baseline methods (MLP, Attention) on all OOD terrains. The largest difference is shown in the pallets terrain where the MLP and Attention methods have a success rate lower than 20\%. Our method achieved around 70\% success rate which is a 3.5 fold improvement.  

\subsubsection{Success rate on stones}
Fig. \ref{fig:quantitative_results}-(b) shows the success rate of each method when evaluated on the stepping stones terrain at all difficulty levels. We first observe that the MLP policy has the overall lowest success rate on all difficulties. the performance drops drastically around level 5 and reaches 0 at level 8. The Attention policy shows high performance up until level 6 but decreases rapidly from levels 7 and above. Our proposed method shows comparable performance to the Attention policy up until level 6. From level 7 and above we observe that our method out performs the baseline policies. At level 8, both the MLP and Attention policies have a success rate near 0\% while our proposed method maintains a higher success rate at around 80\%. This 80\% difference shows that our proposed method especially thrives with high difficulty stepping stones that are smaller and further apart needing more precise foot placement. 

\begin{figure}[tb]
    \centering
    \large
    \includegraphics[width=\columnwidth]{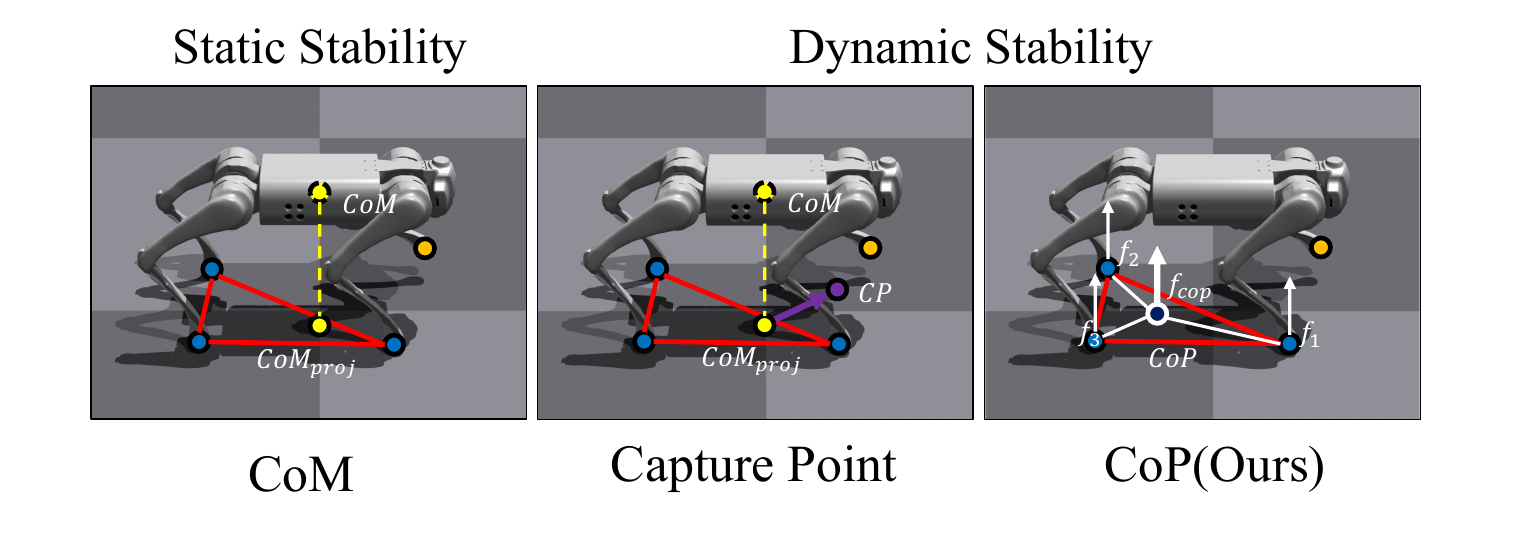}
    \caption{We implement stability rewards based on the CoM and the Capture Point for comparing with our proposed CoP-based stability reward.}
    \label{fig:com_cp_cop}
    \vspace{-10pt}
\end{figure}

\subsubsection{Success rate on stones at different velocities}
In Fig. \ref{fig:quantitative_results}-(c) we present the success rate evaluation on the stones environments at velocities from 0.2 m/s to 1.0 m/s. From Fig. \ref{fig:quantitative_results}-(c) we find that our proposed method out performs the MLP and the Attention policies especially at velocities around 0.3 to 0.5 m/s. At 0.3 m/s we observe a near 80\% improvement over the baseline policies. 

\subsection{Comparison of Heightmap Encoders}
We compare different network architectures for heightmap encoding. 
We present the overall success rate in Fig. \ref{tab:overall_performance}. The overall success rate is calculated from the average success rates of all terrains and all difficulties from 0 to 9. From Fig. \ref{tab:overall_performance} we observe that Our proposed method produces the highest success rates overall compared to baseline methods such as Attention  policies and MLP policies.

\subsection{Qualitative Locomotion Results}
We present qualitative locomotion results in Fig. \ref{fig:qualitative_locomotion_results}. We found that our method enables locomotion on terreains such as stones + stairs up terrain \ref{fig:qualitative_locomotion_results}-(a) and pallets~\ref{fig:qualitative_locomotion_results}-(f) which were not seen during training. Locomotion on other in-domain terrains~\ref{fig:qualitative_locomotion_results}-(b), ~\ref{fig:qualitative_locomotion_results}-(c) and out-of-domain terrains~~\ref{fig:qualitative_locomotion_results}-(d), ~\ref{fig:qualitative_locomotion_results}-(e) are presented as well. 

Fig.~\ref{fig:locomotion_result_stones_ours} depicts a locomotion trajectory with our policy traversing over the stones terrain. From the Contact state plot we observe that the policy adapts to the environment successfully, changing from a standard trot to a leap when overcoming the gap. Interestingly, we observed that the robot slows down right before a gap and accelerates as it overcomes the terrain.

\section{Ablation Studies}

\subsection{Comparison of Stability Rewards}
Table \ref{tab:ablation_stability_rewards} shows the comparison of different stability rewards. We compare a static stability reward based on the CoM and two dynamic stability rewards based on the CP(:Capture Point) and CoP(Ours) as depicted in~\ref{fig:com_cp_cop}. From our results we observe that the CoM stability reward and the CP stability reward decrease the success rate while the CoP stability reward increases the success rate when evaluated on all terrains at a set command velocity of 1.0m/s. In preliminary experiments, the CoM-based static stability reward worked well for terrain that is relatively flat, improving the success rate at lower velocities. However, when the terrains are more complex, the performance degraded. For the CP-based stability reward, we observed a sharper decrease in the success rate. During qualitative evaluation we observed that when the robot is moving at higher speeds (1.0m/s) the CP tends to move out of reach of the support polygon. Since there is a time variable that is not considered in our present reward function, the degraded performance is to be expected.

\begin{table}[t]
\footnotesize
\centering
\caption{Ablation of stability rewards. We found that the CoP reward (Ours) increases the success rate in contrast to Com and CP-based methods.}
\label{tab:ablation_stability_rewards}
\centering
\renewcommand{\arraystretch}{1.5}
\begin{tabular}{r|cccc}
\hline
\textbf{Stability Reward} & None & CoM & CP & CoP(Ours) \\
\hline
\textbf{Success Rate} [\%]~$\uparrow$ & \underline{74.6} & 73.7 & 66.8 & \textbf{77.0} \\
\textbf{Survival Rate} [\%]~$\uparrow$ & \underline{75.3} & 74.6 & 68.6 & \textbf{78.5} \\
\hline
\end{tabular}
\end{table}

\subsection{Comparison of Proposed Components}
In the below Table \ref{tab:ablation_proposed_components} we present the ablation results of our proposed method. To determine which parts of our method had the most impact we test the results of only using the foot position map and both the foot position map and the stability rewards. First, we observe that utilizing the foot position map has the highest increase in success rate at around 10\% jump from the baseline Attention method.  When we utilize the stability rewards we see a 2.4\% increase in success rate as well. From these experiments we conclude that the explicit foot position map has the highest impact on success rate with the stability rewards following.

\begin{table}[tb]
\footnotesize
\centering
\caption{Ablation of proposed components. The foot position map produces 10\% jump and the dynamic stability rewards produce an additional 2.4\% increase in success rates.}
\label{tab:ablation_proposed_components}
\centering
\setlength{\tabcolsep}{3pt}  
\renewcommand{\arraystretch}{1.5}
\begin{tabular}{r|P{4em}P{4em}P{4em}P{4em}}
\hline
\textbf{Component} & Attention & Footmap & Ours \\
\hline
\textbf{Success Rate} [\%]~$\uparrow$ & 64.2 &  \underline{74.6} & \textbf{77.0} \\
\textbf{Survival Rate} [\%]~$\uparrow$ & 68.7 &  \underline{75.4} & \textbf{78.5} \\
\hline
\end{tabular}
\end{table}

\subsection{The Effect of Global Velocity Tracking}
When using the standard local velocity tracking, we found that in some cases the robots learns to avoid difficult terrain like stepping stones. With local velocity tracking, the robot can obtain a relatively high reward without the risk of walking over dangerous terrain by simply turning in a different direction. Since this method does not take into account the absolute direction that the robot is moving towards, the robot can exploit the setup. 

\begin{figure}[t]
    \centering
    \large
    \begin{subfigure}{0.49\columnwidth}
        \centering
        \includegraphics[width=\textwidth]{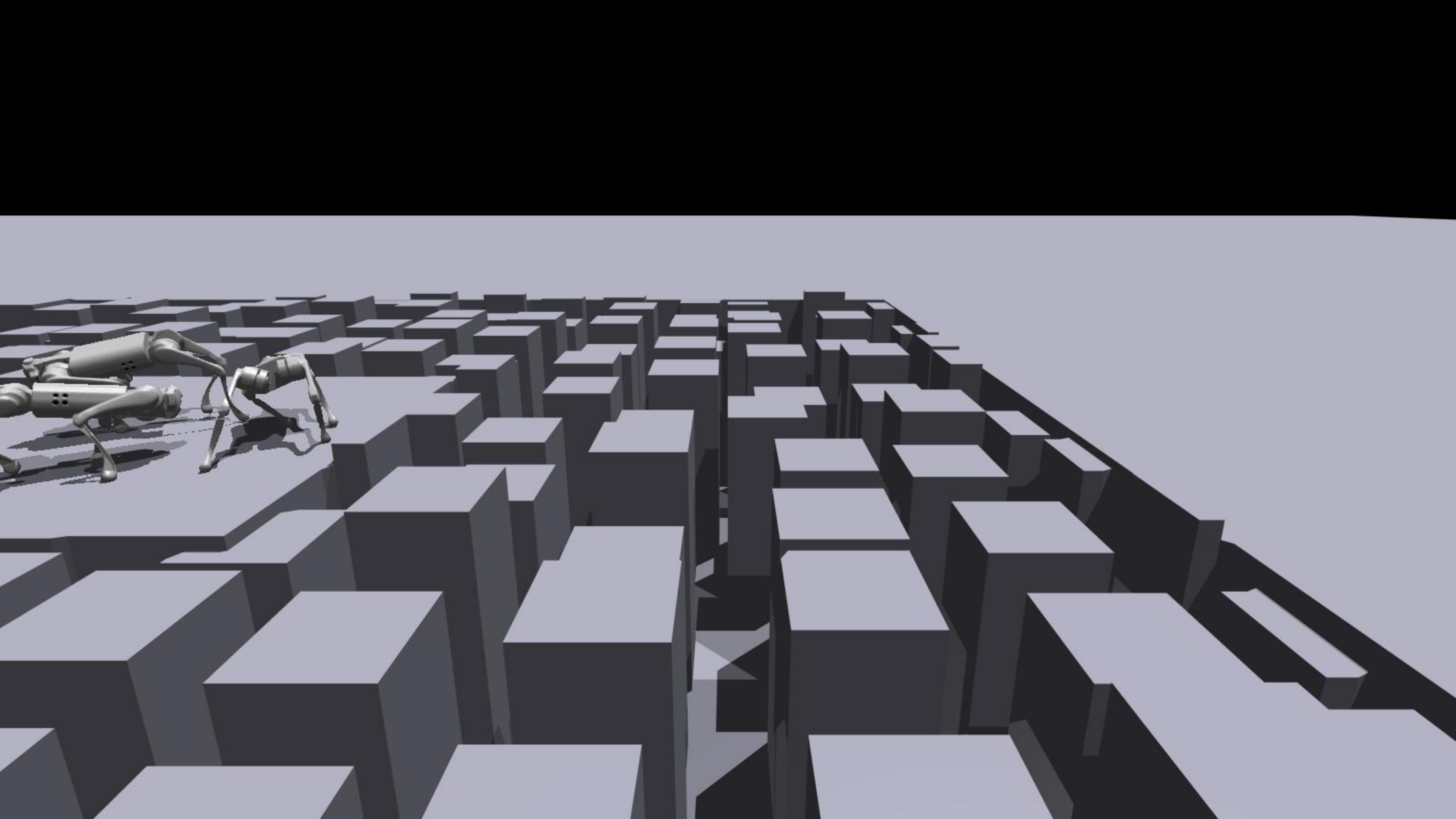}
        \caption{Local velocity tracking}
        \label{fig:footmap_eval_no_global_vel_tracking}
    \end{subfigure}
    \begin{subfigure}{0.49\columnwidth}
        \centering
        \includegraphics[width=\textwidth]{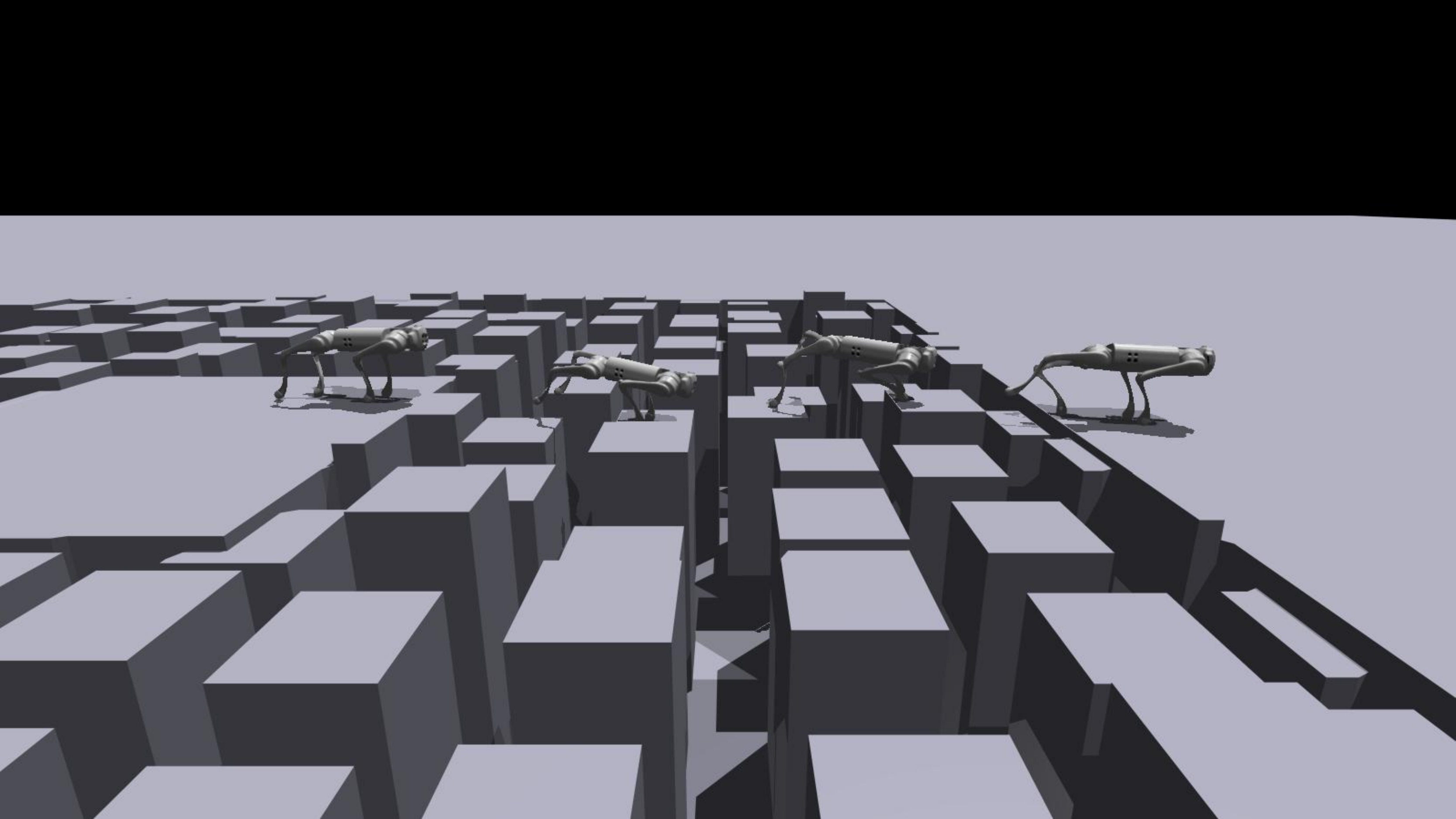}
        \caption{Global velocity tracking}
        \label{fig:footmap_eval_global_vel_tracking}
    \end{subfigure}
    \caption{When using global velocity tracking the issue of robots avoiding difficult terrain is resolved.}
    \label{fig:footmap_qualitative_results_global_vel_tracking}
\end{figure}

To prevent the robot from exploiting the reward function we proposed global velocity tracking. From Fig.~\ref{fig:footmap_eval_global_vel_tracking} we observe that the robot does not avoid the dangerous terrain. 

\section{SIM-TO-SIM TRANSFER EXPERIMENTS}

\begin{figure}[tb]
    \centering
        \centering
        \includegraphics[width=\columnwidth]{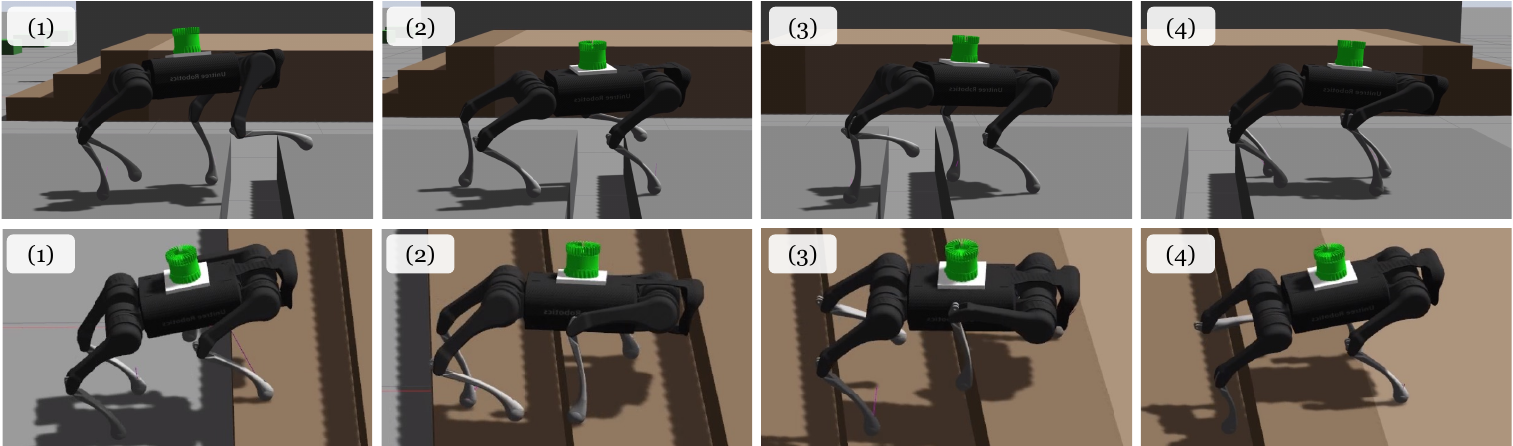}
    \caption{Locomotion results on gaps (top) of 20cm and stairs (bottom) with heights of 20cm. Experiments were conducted in the Gazebo simulator.}
    \label{fig:sim2sim_results}
    \vspace{-10pt}
\end{figure}

We conduct sim-to-sim transfer experiments to validate the applicability of our proposed method. We utilize a Unitree A1 robot equipped with an Ouster OS1-128 Lidar sensor deployed in the Gazsebo simulator. For the heightmap generation we utilize the elevation mapping~\cite{miki2022elevation} ROS package. 
For the sim-to-sim transfer we add noise and domain randomization during the finetuning stage of training. The domain randomization parameters are given in \ref{tab:drand}. As shown in Fig.~\ref{fig:sim2sim_results} we report successful locomotion over gap terrains (top) and stairs terrains (bottom). 


\section{CONCLUSION AND FUTURE WORK}
In this paper we proposed a foot position map, dynamic stability rewards and global velocity tracking to improve locomotion on complex terrain. We demonstrated the capabilities of our policy with extensive tests on in-domain and OOD terrains as well as successful sim-to-sim transfer.

Our future work will include sim-to-real transfer experiments, learning safety limits of locomotion to prevent failure in real world scenarios and further improvements in the encoder architecture.


\addtolength{\textheight}{-12cm}   



\section*{APPENDIX}
\subsection{Training Parameters}

The domain randomization parameters are presented in Table~\ref{tab:drand}, and the PPO training parameters are in Table~\ref{tab:ppo_params}. 

\begin{table}[ht]
    \centering
    \caption{Domain Randomization}
    \begin{tabular}{rcc}
    \toprule 
     \textbf{Parameters} & \textbf{Range} & \textbf{Unit}  \\
     \midrule
     Ground Friction&[0.5,1.25]&-\\
     Ground Restitution&[0.0,0.8]&-\\
     Link Mass &[0.9,1.1]×nominal value&Kg\\
     Payload Mass&[-1,2]&Kg\\
     CoM&[-0.05,0.05]$^3$&m\\
     Motor Strength&[0.9,1.1]×motor torque&Nm\\
     Joint $K_p$&[0.9,1.1]×40&-\\
     Joint $K_d$&[0.9,1.1]×1.0&-\\
     Initial Joint Positions&[0.5,1.5]×default angles&rad\\
     System Delay&[0,40]&ms\\
     External Force&[-30,30]$^3$&N\\
     Heightmap Drift&[-0.05,0.05]$^3$&m\\
     \bottomrule
    \end{tabular}
    \label{tab:drand}
\end{table}

\begin{table}[ht]
    \centering
    \caption{PPO Training Parameters}
    \begin{tabular}{r|c}
    \toprule 
     Hyperparameter&Value  \\
     \midrule
    Clip range&0.2 \\
    Entropy coefficient&0.005(stage1), 0.002(stage2) \\
    Discount factor&0.99 \\
    GAE discount factor&0.95 \\
    Desired KL-divergence&0.01 \\
    Learning rate&1e-4 \\
    Adam epsilon&1e-8 \\
    Replay Buffer Size&4096×24 \\
     \bottomrule
    \end{tabular}
    \label{tab:ppo_params}
\end{table}



\addtolength{\textheight}{12cm}
\bibliographystyle{IEEEtran}
\bibliography{references} 

@ARTICLE{9134750,
  author={Jenelten, Fabian and Miki, Takahiro and Vijayan, Aravind E and Bjelonic, Marko and Hutter, Marco},
  journal={IEEE Robotics and Automation Letters}, 
  title={Perceptive Locomotion in Rough Terrain – Online Foothold Optimization}, 
  year={2020},
  volume={5},
  number={4},
  pages={5370-5376},
  doi={10.1109/LRA.2020.3007427}}

@article{fahmi2022vital,
  title={ViTAL: Vision-Based Terrain-Aware Locomotion for Legged Robots},
  author={Fahmi, Shamel and Barasuol, Victor and Esteban, Domingo and Villarreal, Octavio and Semini, Claudio},
  journal={IEEE Transactions on Robotics},
  year={2022},
  publisher={IEEE}
}

@misc{vaswani2023attentionneed,
      title={Attention Is All You Need}, 
      author={Ashish Vaswani and Noam Shazeer and Niki Parmar and Jakob Uszkoreit and Llion Jones and Aidan N. Gomez and Lukasz Kaiser and Illia Polosukhin},
      year={2023},
      eprint={1706.03762},
      archivePrefix={arXiv},
      primaryClass={cs.CL},
      url={https://arxiv.org/abs/1706.03762}, 
}

@article{Shkolnik2011little_dog,
    author = {Alexander Shkolnik and Michael Levashov and Ian R. Manchester and Russ Tedrake},
    title ={Bounding on rough terrain with the LittleDog robot},
    journal = {The International Journal of Robotics Research},
    volume = {30},
    number = {2},
    pages = {192-215},
    year = {2011},
    doi = {10.1177/0278364910388315},
    URL = {https://doi.org/10.1177/0278364910388315},
    eprint = {https://doi.org/10.1177/0278364910388315},
}

@inproceedings{Grandia2019FeedbackMPC,
  author       = {Ruben Grandia and
                  Farbod Farshidian and
                  Ren{\'{e}} Ranftl and
                  Marco Hutter},
  title        = {Feedback {MPC} for Torque-Controlled Legged Robots},
  booktitle    = {2019 {IEEE/RSJ} International Conference on Intelligent Robots and
                  Systems, {IROS} 2019, Macau, SAR, China, November 3-8, 2019},
  pages        = {4730--4737},
  publisher    = {{IEEE}},
  year         = {2019},
  url          = {https://doi.org/10.1109/IROS40897.2019.8968251},
  doi          = {10.1109/IROS40897.2019.8968251},
}

@ARTICLE{Ruben2022perceptiveNMPC,
  author={Grandia, Ruben and Jenelten, Fabian and Yang, Shaohui and Farshidian, Farbod and Hutter, Marco},
  journal={IEEE Transactions on Robotics}, 
  title={Perceptive Locomotion Through Nonlinear Model-Predictive Control}, 
  year={2023},
  volume={},
  number={},
  pages={1-20},
  doi={10.1109/TRO.2023.3275384}}

@INPROCEEDINGS{9981198,
  author={Rudin, Nikita and Hoeller, David and Bjelonic, Marko and Hutter, Marco},
  booktitle={2022 IEEE/RSJ International Conference on Intelligent Robots and Systems (IROS)}, 
  title={Advanced Skills by Learning Locomotion and Local Navigation End-to-End}, 
  year={2022},
  volume={},
  number={},
  pages={2497-2503},
  doi={10.1109/IROS47612.2022.9981198}}

@ARTICLE{11196002,
  author={Dong, Yinzhao and Ma, Ji and Zhao, Liu and Li, Wanyue and Lu, Peng},
  journal={IEEE Transactions on Robotics}, 
  title={MARG: MAstering Risky Gap Terrains for Legged Robots With Elevation Mapping}, 
  year={2025},
  volume={41},
  number={},
  pages={6123-6139},
  doi={10.1109/TRO.2025.3619041}}

@INPROCEEDINGS{8460731,
  author={Fankhauser, Peter and Bjelonic, Marko and Dario Bellicoso, C. and Miki, Takahiro and Hutter, Marco},
  booktitle={2018 IEEE International Conference on Robotics and Automation (ICRA)}, 
  title={Robust Rough-Terrain Locomotion with a Quadrupedal Robot}, 
  year={2018},
  volume={},
  number={},
  pages={5761-5768},
  doi={10.1109/ICRA.2018.8460731}}

@ARTICLE{10305254,
  author={Luo, Zeren and Xiao, Erdong and Lu, Peng},
  journal={IEEE Robotics and Automation Letters}, 
  title={FT-Net: Learning Failure Recovery and Fault-Tolerant Locomotion for Quadruped Robots}, 
  year={2023},
  volume={8},
  number={12},
  pages={8414-8421},
  doi={10.1109/LRA.2023.3329766}}

@article{Ding_2021,
   title={Representation-Free Model Predictive Control for Dynamic Motions in Quadrupeds},
   volume={37},
   ISSN={1941-0468},
   url={http://dx.doi.org/10.1109/TRO.2020.3046415},
   DOI={10.1109/tro.2020.3046415},
   number={4},
   journal={IEEE Transactions on Robotics},
   publisher={Institute of Electrical and Electronics Engineers (IEEE)},
   author={Ding, Yanran and Pandala, Abhishek and Li, Chuanzheng and Shin, Young-Ha and Park, Hae-Won},
   year={2021},
}

@article{haarnoja2018learning,
  title={Learning to walk via deep reinforcement learning},
  author={Haarnoja, Tuomas and Ha, Sehoon and Zhou, Aurick and Tan, Jie and Tucker, George and Levine, Sergey},
  journal={arXiv preprint arXiv:1812.11103},
  year={2018}
}

@article{Lee_2020,
   title={Learning quadrupedal locomotion over challenging terrain},
   volume={5},
   ISSN={2470-9476},
   url={http://dx.doi.org/10.1126/scirobotics.abc5986},
   DOI={10.1126/scirobotics.abc5986},
   number={47},
   journal={Science Robotics},
   publisher={American Association for the Advancement of Science (AAAS)},
   author={Lee, Joonho and Hwangbo, Jemin and Wellhausen, Lorenz and Koltun, Vladlen and Hutter, Marco},
   year={2020},
   month=oct }

@misc{tsounis2020deepgaitplanningcontrolquadrupedal,
      title={DeepGait: Planning and Control of Quadrupedal Gaits using Deep Reinforcement Learning}, 
      author={Vassilios Tsounis and Mitja Alge and Joonho Lee and Farbod Farshidian and Marco Hutter},
      year={2020},
      eprint={1909.08399},
      archivePrefix={arXiv},
      primaryClass={cs.RO},
      url={https://arxiv.org/abs/1909.08399}, 
}

@article{Miki_2022,
   title={Learning robust perceptive locomotion for quadrupedal robots in the wild},
   volume={7},
   ISSN={2470-9476},
   url={http://dx.doi.org/10.1126/scirobotics.abk2822},
   DOI={10.1126/scirobotics.abk2822},
   number={62},
   journal={Science Robotics},
   publisher={American Association for the Advancement of Science (AAAS)},
   author={Miki, Takahiro and Lee, Joonho and Hwangbo, Jemin and Wellhausen, Lorenz and Koltun, Vladlen and Hutter, Marco},
   year={2022},
   month=jan }

@inproceedings{rudin2022learning,
  title={Learning to walk in minutes using massively parallel deep reinforcement learning},
  author={Rudin, Nikita and Hoeller, David and Reist, Philipp and Hutter, Marco},
  booktitle={Proc. Conference on Robot Learning ({CoRL})},
  pages={91--100},
  year={2022}
}

@article{Jenelten_2024,
   title={DTC: Deep Tracking Control},
   volume={9},
   ISSN={2470-9476},
   url={http://dx.doi.org/10.1126/scirobotics.adh5401},
   DOI={10.1126/scirobotics.adh5401},
   number={86},
   journal={Science Robotics},
   publisher={American Association for the Advancement of Science (AAAS)},
   author={Jenelten, Fabian and He, Junzhe and Farshidian, Farbod and Hutter, Marco},
   year={2024},
   month=jan }

@article{Jenelten_2022,
   title={TAMOLS: Terrain-Aware Motion Optimization for Legged Systems},
   volume={38},
   ISSN={1941-0468},
   url={http://dx.doi.org/10.1109/TRO.2022.3186804},
   DOI={10.1109/tro.2022.3186804},
   number={6},
   journal={IEEE Transactions on Robotics},
   publisher={Institute of Electrical and Electronics Engineers (IEEE)},
   author={Jenelten, Fabian and Grandia, Ruben and Farshidian, Farbod and Hutter, Marco},
   year={2022},
   month=dec, pages={3395–3413} }

@misc{zhang2024learningagilelocomotionrisky,
      title={Learning Agile Locomotion on Risky Terrains}, 
      author={Chong Zhang and Nikita Rudin and David Hoeller and Marco Hutter},
      year={2024},
      eprint={2311.10484},
      archivePrefix={arXiv},
      primaryClass={cs.RO},
      url={https://arxiv.org/abs/2311.10484}, 
}

@misc{yu2025walkingterrainreconstructionlearning,
      title={Walking with Terrain Reconstruction: Learning to Traverse Risky Sparse Footholds}, 
      author={Ruiqi Yu and Qianshi Wang and Yizhen Wang and Zhicheng Wang and Jun Wu and Qiuguo Zhu},
      year={2025},
      eprint={2409.15692},
      archivePrefix={arXiv},
      primaryClass={cs.RO},
      url={https://arxiv.org/abs/2409.15692}, 
}

@misc{wang2025beamdojolearningagilehumanoid,
      title={BeamDojo: Learning Agile Humanoid Locomotion on Sparse Footholds}, 
      author={Huayi Wang and Zirui Wang and Junli Ren and Qingwei Ben and Tao Huang and Weinan Zhang and Jiangmiao Pang},
      year={2025},
      eprint={2502.10363},
      archivePrefix={arXiv},
      primaryClass={cs.RO},
      url={https://arxiv.org/abs/2502.10363}, 
}

@article{PPO,
  title={Proximal policy optimization algorithms},
  author={Schulman, John and Wolski, Filip and Dhariwal, Prafulla and Radford, Alec and Klimov, Oleg},
  journal={arXiv preprint arXiv:1707.06347},
  year={2017}
}

@Article{s20174911,
AUTHOR = {Hao, Qian and Wang, Zhaoba and Wang, Junzheng and Chen, Guangrong},
TITLE = {Stability-Guaranteed and High Terrain Adaptability Static Gait for Quadruped Robots},
JOURNAL = {Sensors},
VOLUME = {20},
YEAR = {2020},
NUMBER = {17},
ARTICLE-NUMBER = {4911},
URL = {https://www.mdpi.com/1424-8220/20/17/4911},
PubMedID = {32878028},
ISSN = {1424-8220},
DOI = {10.3390/s20174911}
}

@INPROCEEDINGS{10611621,
  author={Duan, Helei and Pandit, Bikram and Gadde, Mohitvishnu S. and Van Marum, Bart and Dao, Jeremy and Kim, Chanho and Fern, Alan},
  booktitle={2024 IEEE International Conference on Robotics and Automation (ICRA)}, 
  title={Learning Vision-Based Bipedal Locomotion for Challenging Terrain}, 
  year={2024},
  volume={},
  number={},
  pages={56-62},
  doi={10.1109/ICRA57147.2024.10611621}}

@inproceedings{miki2022elevation,
  title={Elevation mapping for locomotion and navigation using gpu},
  author={Miki, Takahiro and Wellhausen, Lorenz and Grandia, Ruben and Jenelten, Fabian and Homberger, Timon and Hutter, Marco},
  booktitle={2022 IEEE/RSJ International Conference on Intelligent Robots and Systems (IROS)},
  pages={2273--2280},
  year={2022},
  organization={IEEE}
}

@article{
doi:10.1126/scirobotics.adv3604,
author = {Junzhe He  and Chong Zhang  and Fabian Jenelten  and Ruben Grandia  and Moritz Bächer  and Marco Hutter },
title = {Attention-based map encoding for learning generalized legged locomotion},
journal = {Science Robotics},
volume = {10},
number = {105},
pages = {eadv3604},
year = {2025},
doi = {10.1126/scirobotics.adv3604},
URL = {https://www.science.org/doi/abs/10.1126/scirobotics.adv3604},
eprint = {https://www.science.org/doi/pdf/10.1126/scirobotics.adv3604}}

@inproceedings{
makoviychuk2021isaac,
title={Isaac Gym: High Performance {GPU} Based Physics Simulation For Robot Learning},
author={Viktor Makoviychuk and Lukasz Wawrzyniak and Yunrong Guo and Michelle Lu and Kier Storey and Miles Macklin and David Hoeller and Nikita Rudin and Arthur Allshire and Ankur Handa and Gavriel State},
booktitle={Thirty-fifth Conference on Neural Information Processing Systems Datasets and Benchmarks Track (Round 2)},
year={2021},
}

@inproceedings{
  yang2022learning,
  title={Learning Vision-Guided Quadrupedal Locomotion End-to-End with Cross-Modal Transformers},
  author={Ruihan Yang and Minghao Zhang and Nicklas Hansen and Huazhe Xu and Xiaolong Wang},
  booktitle={International Conference on Learning Representations},
  year={2022},
  url={https://openreview.net/forum?id=nhnJ3oo6AB}
}

@article{Xiao2025,
  author  = {Xiao, Erdong and Dong, Yinzhao and Lam, James and Lu, Peng},
  title   = {Learning stable bipedal locomotion skills for quadrupedal robots on challenging terrains with automatic fall recovery},
  journal = {npj Robotics},
  year    = {2025},
  volume  = {3},
  number  = {1},
  pages   = {22},
  doi     = {10.1038/s44182-025-00043-2},
  url     = {https://doi.org/10.1038/s44182-025-00043-2},
  issn    = {2731-4278}
}

@INPROCEEDINGS{4115602,
  author={Pratt, Jerry and Carff, John and Drakunov, Sergey and Goswami, Ambarish},
  booktitle={2006 6th IEEE-RAS International Conference on Humanoid Robots}, 
  title={Capture Point: A Step toward Humanoid Push Recovery}, 
  year={2006},
  volume={},
  number={},
  pages={200-207},
  doi={10.1109/ICHR.2006.321385}}

@ARTICLE{9779429,
  author={Gangapurwala, Siddhant and Geisert, Mathieu and Orsolino, Romeo and Fallon, Maurice and Havoutis, Ioannis},
  journal={IEEE Transactions on Robotics}, 
  title={RLOC: Terrain-Aware Legged Locomotion Using Reinforcement Learning and Optimal Control}, 
  year={2022},
  volume={38},
  number={5},
  pages={2908-2927},
  doi={10.1109/TRO.2022.3172469}}

@InProceedings{10.1007/978-3-031-21090-7_31,
author="Xie, Zhaoming
and Da, Xingye
and Babich, Buck
and Garg, Animesh
and de Panne, Michiel van",
editor="LaValle, Steven M.
and O'Kane, Jason M.
and Otte, Michael
and Sadigh, Dorsa
and Tokekar, Pratap",
title="GLiDE: Generalizable Quadrupedal Locomotion in Diverse Environments with a Centroidal Model",
booktitle="Algorithmic Foundations of Robotics XV",
year="2023",
publisher="Springer International Publishing",
address="Cham",
pages="523--539",
isbn="978-3-031-21090-7"
}

@article{doi:10.1142/S0219843604000083,
author = {VUKOBRATOVIĆ, MIOMIR and BOROVAC, BRANISLAV},
title = {ZERO-MOMENT POINT — THIRTY FIVE YEARS OF ITS LIFE},
journal = {International Journal of Humanoid Robotics},
volume = {01},
number = {01},
pages = {157-173},
year = {2004},
doi = {10.1142/S0219843604000083},

URL = { 
    
        https://doi.org/10.1142/S0219843604000083
    }
}

\end{document}